\documentclass[a4paper,UKenglish,cleveref,autoref, thm-restate]{lipics-v2021}
\usepackage[utf8]{inputenc}
\usepackage{amsfonts}
\usepackage{amsmath}
\usepackage{amsthm,stmaryrd,wrapfig}
\usepackage[ruled,vlined,linesnumbered,noend]{algorithm2e}
\newtheorem{sublemma}{Sublemma}
\usepackage{subcaption}

\newcommand{\introparagraph}[1]{\vspace{0.7mm} \noindent \textbf{\em #1.}}

\newcommand{\frep}[1]{\textsc{Forbidden-Repair}\langle{#1} \rangle}
\newcommand{\dom}[0]{\mathbb{D}}
\newcommand{\key}[1]{\mathsf{key}(#1)}
\usepackage{xcolor}

\usepackage{tikz-cd}
\usetikzlibrary{automata,positioning,arrows,shapes,decorations.pathmorphing}

\title{Certifiable Robustness for Nearest Neighbor Classifiers} 

\author{Austen Z. {Fan}}{Department of Computer Science, University of Wisconsin-Madison, United States}{afan@cs.wisc.edu}{https://orcid.org/0000-0001-7714-2195}{}

\author{Paraschos {Koutris}}{Department of Computer Science, University of Wisconsin-Madison, United States \and \url{http://pages.cs.wisc.edu/~paris/}}{koutris@wisc.edu}{https://orcid.org/0000-0001-6309-1702}{}

\authorrunning{A.\,Z. Fan and P. Koutris} 

\Copyright{Austen Z. Fan and Paraschos Koutris} 

\ccsdesc{Theory of computation~Database theory}
\ccsdesc{Theory of computation~Incomplete, inconsistent, and uncertain databases}

\keywords{Inconsistent databases, $k$-NN classification, certifiable robustness} 

\funding{This research was supported in part by National Science Foundation grants CRII-1850348 and III-1910014, as well as a gift by Google.}

\nolinenumbers 

\EventEditors{Dan Olteanu and Nils Vortmeier}
\EventNoEds{2}
\EventLongTitle{25th International Conference on Database Theory (ICDT 2022)}
\EventShortTitle{ICDT 2022}
\EventAcronym{ICDT}
\EventYear{2022}
\EventDate{March 29--April 1, 2022}
\EventLocation{Edinburgh, UK}
\EventLogo{}
\SeriesVolume{220}
\ArticleNo{18}

\begin{document}

\maketitle

\begin{abstract}
ML models are typically trained using large datasets of high quality. However, training datasets often contain inconsistent or incomplete data. To tackle this issue, one solution is to develop algorithms that can check whether a prediction of a model is {\em certifiably robust}. Given a learning algorithm that produces
a classifier and given an example at test time, a classification outcome is certifiably robust if it is predicted by every model trained across all possible worlds (repairs) of the uncertain (inconsistent) dataset. This notion of robustness falls naturally under the framework of certain answers.
In this paper, we study the complexity of certifying robustness for a simple but widely deployed classification algorithm, $k$-Nearest Neighbors ($k$-NN). Our main focus is on inconsistent datasets when the integrity constraints are functional dependencies (FDs). For this setting, we establish a dichotomy in the complexity of certifying robustness w.r.t.\ the set of FDs: the problem either admits a polynomial time algorithm, or it is \textsf{coNP-hard}. Additionally, we exhibit a similar dichotomy for the counting version of the problem, where the goal is to count the number of possible worlds that predict a certain label. As a byproduct of our study, we also establish the complexity of a problem related to finding an optimal subset repair that may be of independent interest.
\end{abstract}

\section{Introduction}

Machine Learning (ML) has been widely adopted as a central tool in business analytics, medical decisions, autonomous driving, and many other domains. In supervised learning settings, ML models are typically trained using large datasets of high quality. However, real-world training datasets often contain incorrect or incomplete data. For example, attribute values may be missing from the dataset, attribute values may be wrong, or the dataset can violate integrity constraints.
Several approaches to tackle this problem have been proposed in the literature, including data cleaning~\cite{KrishnanWWFG16,HoloClean} and robust learning methods~\cite{DiakonikolasKKL19, SteinhardtKL17, DrewsAD20}.

In this work, we study an alternative but complementary approach using the framework of {\em certain answers}, which has been a focus of the database community in the last two decades~\cite{ArenasBC99,Chomicki07}. Under this framework, an inconsistent or incomplete training dataset is modeled as a set of possible worlds called {\em repairs}. We can think of a repair as a way to clean the dataset such that it is consistent (w.r.t.\ a set of integrity constraints) and complete. In the context of query answering, certain answers are the output tuples that exist in the query result of every possible world. In other words, we will obtain a certain answer in the output no matter how we choose to repair the dataset.

The notion of certain answers in the context of ML is known as certifiable (or certified) robustness~\cite{CohenRK19,SteinhardtKL17}.
Given a learning algorithm that produces a classifier and a tuple at test time, we say that a predicted label is {\em certifiably robust} if it is predicted by every model trained across all possible worlds. In other words, certifiably robust predictions come with a formal guarantee that the label is robust to how the training dataset is fixed.
Such a guarantee can be beneficial to decide whether we should trust the predictions of the ML model or whether we should spend resources to clean the dataset before feeding it to the learning algorithm. 

As a first step towards certifying robustness for other more complex ML algorithms, we focus in this work on $k$-Nearest Neighbor classification ($k$-NN). In this problem, we start with a $d$-dimensional dataset equipped with a distance function. Given a test point $x$, the classifier first finds the $k$ points closest to $x$ w.r.t.\ the given distance function, and assigns to $x$ the label that is a plurality among these. Certified robustness for $k$-NNs was recently explored by Karlas et.\ al~\cite{KarlasLWGC0020} under the uncertain model where tuples (training points) with the same label are grouped into blocks, and a possible world is constructed by picking independently exactly one tuple from each block. This setting is equivalent to considering the subset repairs of an inconsistent dataset where the only integrity constraint is a primary key, with the additional restriction that training points with the same key must have the same label. 
\begin{wraptable}{l}{5cm}
 \begin{tabular}{c | c c c |c} 
  & A & B & C & label \\
 \hline\hline
 $t_1$ & 1 & 0 & a & 0 \\ 
 \hline
 $t_2$ & 1 & 2 & b & 0 \\
 \hline
 $t_3$ & 2 & 0 & a & 2 \\
 \hline
 $t_4$ & 2 & 5 & c & 1 \\
 \hline
 $t_5$ & 3 & 1 & a & 0 \\ 
 \hline
  $t_6$ & 4 & 2 & d & 2 \\ 
\end{tabular}
\label{tab}
\end{wraptable}

In this paper, we generalize the study of certifying robustness for $k$-NNs to other inconsistent and uncertain models. Specifically, we consider subset repairs of a dataset where the integrity constraints can be any set of functional dependencies (FDs) and not only a primary key as in~\cite{KarlasLWGC0020}. 

\begin{example}
Consider the inconsistent dataset in the table, where the integrity constraint is the FD $A \rightarrow B$. 
Let the distance function between two tuples $s,t$ be $f(s,t) = |s[A]-t[A]| + |s[B]- t[B]|$.
Suppose that we want to label the test point $x = (0,0,d)$ using a 3-NN classifier. 
This induces the following ordering of the tuples w.r.t.\ their distance from $x$ (for convenience, we include the labels as well):
$$ t_1:\mathbf{0} < t_3:\mathbf{2} < t_2: \mathbf{0} < t_5:\mathbf{0} < t_6:\mathbf{2} < t_4:\mathbf{1}$$ 
A repair for this inconsistent dataset has to choose one tuple from $\{t_1, t_2\}$ and one tuple from $\{t_3, t_4\}$. There are 4 possible repairs, which form the following 3-neighborhoods around $x$:
\begin{align*}
& \{t_1: \mathbf{0}, t_5:\mathbf{0}, t_6: \mathbf{2}\}, & \quad \{t_2: \mathbf{0}, t_5:\mathbf{0}, t_6: \mathbf{2}\} \\
& \{t_1: \mathbf{2}, t_3:\mathbf{0}, t_5: \mathbf{0}\}, & \quad \{t_3: \mathbf{2}, t_2: \mathbf{0}, t_5: \mathbf{0}\}
\end{align*}
In all repairs, label 0 occurs two times, and hence it is always the majority label. Hence, we can certify that $0$ is a robust label for tuple $x$.   
\end{example}

We show that for general sets of FDs the complexity landscape for certified robustness admits a dichotomy: it is computationally intractable for some sets of FDs, and is in polynomial time for the other sets of FDs. We also investigate certifying robustness for other widely used uncertain models, including Codd-tables~\cite{IL84}, or-sets and $?$-sets~\cite{SBHW06}. We establish that in these settings the problem can always be solved in polynomial time.

Our work shows that the logical structure of the errors (or missing values) from a training set can be exploited to construct fast algorithms for certifiable robustness. Tools developed in the database theory community can facilitate the design of these algorithms. We also demonstrate that, even for the relatively simple $k$-NN classifier, the complexity landscape exhibits a complex behavior that is related to other problems in consistent query answering.  

\introparagraph{Our Contribution}
We now present in more detail the contributions of this work:
\begin{itemize}
\item We establish a complexity dichotomy for certifying robustness for $k$-NNs under subset repairs (Section~\ref{sec:main}) into \textsf{P} and \textsf{coNP-complete}. The dichotomy depends on the structure of FDs. More precisely, the syntactic condition for the dichotomy is based on the notion of {\em lhs chains}. In fact, it is the same as the one used for the complexity classification of the problem of counting the number of subset repairs under a set of FDs~\cite{LKW21}. In the case where the only FD constraint is a primary key, we show that we can design an even faster algorithm that runs in linear time in the size of the data, improving upon the running time of the algorithm in~\cite{KarlasLWGC0020}  (Section~\ref{sec:algorithm}).
\item In addition to certified robustness, we study the related problem of {\em counting} the number of repairs that predict a given label for a test point (Section~\ref{sec:counting}). We establish a dichotomy into the complexity classes \textsf{FP} and \textsf{\#P-complete} with the same syntactic condition as the decision problem. The polynomial time algorithm here depends exponentially on the number of classification labels, in contrast to our algorithm for certifiable robustness which has a linear dependence on the number of labels.
\item We show that certifying robustness for $k$-NNs is tightly connected to the problem of finding the {\em subset repair with the smallest total weight}, when each tuple is assigned a positive weight (Section~\ref{sec:forbidden}). As a consequence, we obtain a dichotomy result for that problem as well. Note that this problem is a symmetric variant of the problem in~\cite{LKR18}, which asks instead for the repair with the maximum total weight.

\item Finally, we investigate the complexity of certifiable robustness for $k$-NNs for three widely used uncertain models: Codd-tables, $?$-sets and or-sets (Section~\ref{sec:other}). We show that for all the above models, certifying robustness for $k$-NN classification admits a polynomial time algorithm.
\end{itemize}
\section{Related Work}

\introparagraph{Certain Query Answering} There has been a long line of research in {\em Certain Query Answering} (CQA) in the database community. Data consistency might be violated, for example, during data integration or in a data warehouse. It is then natural to ask: given a query and such inconsistent data, can we answer the query with a certain answer? Arenas, Bertossi, and Chomicki~\cite{ArenasBC99} first define the notion of a repair, which refers to a consistent subinstance that is minimally different from the inconsistent data. A certain answer to the query is defined as an answer that will result from every repair. Beginning from the work of Fuxman and Miller~\cite{FuxmanM07}, more general dichotomy results in CQA have been proven~\cite{KolaitisP12, KoutrisS14, KoutrisW15}. A dichotomy theorem for a class of queries and integrity constraints says CQA is either in polynomial time or \textsf{coNP-complete}, usually depending on an efficiently checkable criterion for tractability. Certain answers have also been studied in the context of incomplete models of data~\cite{Libkin11, RazniewskiN11, 2012Greco}. 

\introparagraph{Subset Repairs} Livshits et.\ al~\cite{LKW21} studied the problem of counting the number of subset repairs w.r.t.\ a given set of FDs, establishing a complexity dichotomy. The syntactic condition for tractability (existence of a lhs chain) is the same as the one we establish for certifiable robustness in $k$-NN classification. Livshits et.\ al~\cite{LKR18} also studied the problem of finding a maximum-weight subset repair w.r.t. a given set of FDs, and showed that the complexity observes a dichotomy. In this paper we study the symmetric problem of finding a minimum-weight subset repair, and show that the problem also exhibits a complexity dichotomy, albeit the condition for tractability is again the existence of a lhs chain.


\introparagraph{Certifiable Robustness in ML}
Robust learning methods are used to learn over inconsistent or incomplete datasets. For example, \cite{DrewsAD20} discusses a sound verification technique which proves whether a prediction is robust to data poisoning in decision-tree models. There is also a line of work on smoothing techniques for ML robustness~\cite{CohenRK19, JiaCWG20, RosenfeldWRK20, Kumar0GF20}, where added random noise, usually with a Gaussian distribution, will sometimes boost the robustness of the model against adversarial attacks such as label-flipping. Our approach is different in that we prove a dichotomy for $k$-NN certifiable robustness, i.e.\ {\em either} we can assert that the dataset will always lead to the same prediction efficiently {\em or} it is \textsf{coNP-complete} to do so, with an efficiently testable criterion. We show how to extend our model to capture scenarios including uncertain labels, weighted tuples, and data poisoning.
\section{Preliminaries}

In this paper, we consider relation schemas of the form $R(A_1, \dots, A_d)$ with arity $d$. The attributes $A_1, \dots, A_d$ take values from a (possibly infinite) domain $\dom$. Given a tuple $t$ in an instance $I$ over $R$, we will use $t[A_i]$ to denote its value at attribute $A_i$. It will be convenient to interpret an instance $I$ as a training set of points in the $d$-dimensional space $\dom^d$. We will use the terms point/tuple interchangeably in the paper. 

An {\em uncertain instance} $\mathcal{I}$ over a schema $R(A_1, \dots, A_d)$ is a set of instances over the schema. We will often refer to each instance in $\mathcal{I}$ as a {\em possible world}. We will see later different ways in which we can define uncertain instances implicitly. 

For each tuple $t$ that occurs in some possible world in $\mathcal{I}$, we associate a {\em label} $L(t)$ which takes values from a finite set $\mathcal{Y}$. We will say that the uncertain instance $\mathcal{I}$ equipped with the labeling function $L$ is a {\em labeled uncertain instance} over the schema $R(A_1, \dots, A_d)$. We similarly define a labeled instance $I$.

\introparagraph{Certifiable Robustness} In this work, we will focus on the classification task. Let $\mathcal{L}$ be a learning algorithm that takes as training set a labeled instance $I$ over the schema $R(A_1, \dots, A_d)$, and returns a {\em classifier}, which is a total function $\mathcal{L}_I: \dom^d \rightarrow \mathcal{Y}$.

\begin{definition}[Certifiable Robustness]
Let $\mathcal{I}$ be a labeled uncertain instance over $R(A_1, \dots, A_d)$ and  $\mathcal{L}$ be a classification learning algorithm with labels in $\mathcal{Y}$. We say that a (test) point $x \in \dom^d$ is {\em certifiably robust} in $\mathcal{I}$ if there exists a label $\ell \in \mathcal{Y}$ such that for every possible world $I \in \mathcal{I}$, $\mathcal{L}_I(x) = \ell$.
The label $\ell$ is called a {\em certain label} for $x$.
\end{definition} 

In other words, suppose we call $\ell$ a {\em possible label} for $x$ if there exists some possible world $I \in \mathcal{I}$ for which $\mathcal{L}_I(x) = \ell$, then certifiable robustness simply means that there is only one possible label for $x$.


\introparagraph{Nearest Neighbor Classifiers} In $k$-NN classification, we are given a labeled instance $I$ over $R(A_1, \dots, A_d)$, along with a distance function $f$ over $\dom^d$. 
Given a point $x \in \dom^d$, the classifier first finds the $k$-neighborhood $\mathcal{N}_k(x,I)$, which consists of the $k$ points closest to $x$ w.r.t.\ the distance function $f$. Then, the classifier assigns to $x$ the label that is a plurality among $\mathcal{N}_k(x,I)$. 
When $k=1$, the classifier returns the label of the point that is closest to $x$ w.r.t.\ the distance function $f$. When $|\mathcal{Y}| = 2$, we are performing binary classification. 
We will also consider the generalization of $k$-NN where each tuple has a positive weight, and the classifier assigns the label with the largest total weight.

For this work, we require the following tie-breaking mechanisms: $(i)$ if there are two labels in $\mathcal{N}_k(x,I)$ with the maximum number, then we say $x$ is not certifiably robust for $k$-NN classification, and $(ii)$ if there are more tuples with the same distance to the test point that will make $\mathcal{N}_k(x,I)$ not well-defined, we will break ties according to a predefined ordering of the tuples in the instance.
By a slight abuse of notation, throughout the proof when we say $\mathcal{L}_I(x) = \ell$, we mean the number of tuples labeled $\ell$ is \emph{strictly} more than that of any other labels for any choices made to pick $\mathcal{N}_k(x,I)$. 

\introparagraph{Functional Dependencies} A functional dependency (FD) is an expression of the form $X \rightarrow Y$, where $X$ and $Y$ are sets of attributes from $R$. An instance $I$ over $R$ satisfies $X \rightarrow Y$ if for every two tuples in $I$, if they agree on $X$ they must also agree on $Y$. We say that $I$ satisfies a set of functional dependencies $\Sigma$ if it satisfies every functional dependency in $\Sigma$. For an attribute $A$ and set of FDs $\Sigma$, we define $\Sigma-A$ to be the FD set where we have removed from any FD in $\Sigma$ the attribute $A$.
An {\em FD schema} $\mathbf{R}$ is a pair $(R(A_1, \dots, A_d), \Sigma)$, where $\Sigma$ is a set of FDs defined over $R$.

\introparagraph{Repairs} Given $\Sigma$, assume that we have an inconsistent instance $D$ that violates the functional dependencies in $\Sigma$. We say that $D'$ is a {\em repair} of $D$ if it is a maximal subset of $D$ that satisfies $\Sigma$. In other words, we can create a repair by removing the smallest possible number of tuples from $D$ such that all the integrity constraints are satisfied. We will use $I_\Sigma(D)$ to denote the set of all possible repairs of $D$ w.r.t.\ the FD schema $\Sigma$. If the instance $D$ is consistent, namely it does not violate any functional dependency, then $I_\Sigma(D)$ is defined to be $D$ itself.

\subsection{Problem Definitions}

Although our algorithms work for any distance function such that $f(x,x')$ can be computed in time $O(1)$ (assuming the dimension $d$ is fixed), for the hardness results we need a more precise formalization. We consider two variants of the problem. In the first variant, we will consider a specific distance function, the $p$-norm when the domain is $\dom = \mathbb{R}$. Recall that for any $p\geq 1$, the $p$-norm is
$$\| x - x'\|_p = \left( \sum_{i=1}^d |x[A_i] - x'[A_i]|^p \right)^{1/p}$$ 

For the following definitions, we fix the dimension $d$ and the label set $\mathcal{Y}$.
Formally, we can now define the following decision problem, parameterized by an FD schema $\mathbf{R}$ and an integer $k > 0$.

\begin{definition}[$\textsc{CR-NN}_{p}\langle \mathbf{R},k \rangle$] Given an inconsistent labeled instance $D$ over an FD schema $\mathbf{R}$ and a test point $x$, is $x$ certifiably robust in $I_\Sigma(D)$ for $k$-NN classification w.r.t.\ the $p$-norm?
\end{definition}

Note that $k$ is fixed in the above problem. We can also define the decision problem where $k$ is instead an input parameter, denoted as $\textsc{CR-NN}_{p}\langle \mathbf{R} \rangle$.

In the second variant of the problem, instead of fixing a distance function, we will directly provide as input to the problem a strict ordering of the points in the dataset $D$ w.r.t.\ their distance from the test point $x$. From an algorithmic point of view this does not make much difference, since we can compute the ordering in time $O(|D| \log |D|)$ for any distance function that can be computed in time $O(1)$.

\begin{definition}[$\textsc{CR-NN}_{<}\langle \mathbf{R},k \rangle$] Given an inconsistent labeled instance $D$ over an FD schema $\mathbf{R}$ and a strict ordering of the points in $D$ w.r.t.\ their distance from a test point $x$, is $x$ certifiably robust in $I_\Sigma(D)$ for $k$-NN classification?
\end{definition}

Similarly we also define the problem $\textsc{CR-NN}_{<}\langle \mathbf{R}\rangle$ with the parameter $k$ as part of the input.
Note here that there is a straightforward many-one polynomial time reduction from $\textsc{CR-NN}_{p}\langle \mathbf{R},k \rangle$ to $\textsc{CR-NN}_{<}\langle \mathbf{R},k \rangle$.


Finally, we define the counting variant of the problem. Given an inconsistent instance $D$ and a label $\ell \in \mathcal{Y}$, we want to count how many repairs of $D$ will predict label $\ell$. 

\begin{definition}[$\textsc{\#CR-NN}_< \langle \mathbf{R} \rangle$]
Given an inconsistent labeled instance $D$ over an FD schema $\mathbf{R}$,  a strict ordering of the points in $D$ w.r.t.\ their distance from a test point $x$, and a label $\ell \in \mathcal{Y}$, output the number of repairs in $I_\Sigma(D)$ for which the $k$-NN classifier assigns label $\ell$ to $x$.
\end{definition}

Similarly, one can define the counting question $\textsc{\#CR-NN}_p \langle \mathbf{R} \rangle$.

\section{Main Results} 
\label{sec:main}

In this section, we present and discuss our key results. The main dichotomy theorem relies on the notion of lhs chains for an FD schema, as defined in~\cite{LKW21}.

\begin{definition}[lhs Chain]
 A set of FDs $\Sigma$ has a left-hand-side chain (lhs chain for short) if for every two FDs $X_1 \rightarrow Y_1$ and $X_2 \rightarrow Y_2$ in $\Sigma$, either $X_1 \subseteq X_2$ or $X_2 \subseteq X_1$.
 \end{definition}

One can determine efficiently whether an FD schema is equivalent to one with an lhs chain or not~\cite{LKW21}. 

\begin{example}
Consider the relational schema $R(A,B,C,D)$. The FD set $\{A \rightarrow C, B \rightarrow C\}$ does not have an lhs-chain, since neither of the two left-hand-sides of the FDs are contained in each other. The FD set $\{A B \rightarrow C, B \rightarrow D\}$ on the other hand has an lhs chain.
\end{example}

We are now ready to state our main theorem.

\begin{theorem}[Main Theorem]
Let $\mathbf{R}$ be an FD schema. Then, the following hold:
\begin{itemize}
\item If $\mathbf{R}$ is equivalent to an FD schema with an lhs chain, then $\textsc{CR-NN}_<\langle \mathbf{R} \rangle$ (and thus $\textsc{CR-NN}_p\langle \mathbf{R} \rangle$) is in \textsf{P}.
\item Otherwise, for any integer $k \geq 1$, $\textsc{CR-NN}_p\langle \mathbf{R}, k \rangle$ (and thus $\textsc{CR-NN}_<\langle \mathbf{R},k \rangle$) is \textsf{coNP-complete}.
\end{itemize}
Moreover, it can be decided in polynomial time which of the two cases holds.
\end{theorem}

We show the polynomial time algorithm in Section~\ref{sec:algorithm} and the hardness proof in Section~\ref{sec:hardness}. We should discuss three things at this point. First, the polynomial time algorithm works for any distance function, as long as we can compute the distance between any two points in time $O(1)$. Indeed, the distance function is only needed to compute the order of tuples in the dataset according to their distance from the test point. Second, we show the intractability result for the $p$-norm distance function, which is widely used in practice. It is likely that the problem remains hard for other popular distance functions as well. Third, as we will see in the next section, the tractable cases are polynomial in $k$, the size of the neighborhood. This is important, since $k$ is often set to be a function of the training set size, for example $\sqrt{n}$. For the intractable cases, the problem is already hard even for $k=1$.

\introparagraph{Uncertain Labels} The above dichotomy theorem holds even when we allow inconsistent labels. We can model inconsistent labels by modifying the labelling function $L(t)$ to take values from $\mathcal{P}(\mathcal{Y})$, the power set of the finite label set $\mathcal{Y}$. The set of possible worlds is then defined to be the set of possible worlds of the inconsistent instance $D$ paired with a labelling function $L'$ such that $L'(t) \in L(t)$ for all $t \in D$. The definition of certifiable robustness carries over to this set of possible worlds.

We can simulate uncertain labels by adding an extra attribute (\textsf{label}) to the schema and an FD $A_1, \dots, A_d \rightarrow \textsf{label}$. It is easy to see that the new schema is equivalent to one with an lhs chain if and only if the original one is. Thus, we conclude that {\em uncertain labels do not change the complexity with respect to certifiable robustness.}

\introparagraph{Counting} For the counting variant of certifying robustness for $k$-NNs, we show an analogous dichotomy result. 

\begin{theorem}\label{ref:counting}
Let $\mathbf{R}$ be an FD schema. Then, the following hold:
\begin{itemize}
\item If $\mathbf{R}$ is equivalent to an FD schema with an lhs chain, then $\textsc{\#CR-NN}_<\langle \mathbf{R} \rangle$ is in \textsf{FP}.
\item Otherwise,  even $\textsc{\#CR-NN}_<\langle \mathbf{R}, 1 \rangle$ is \textsf{\#P-complete}.
\end{itemize}
Moreover, it can be decided in \textsf{P} which of the two cases holds.
\end{theorem}

\section{Tractable Cases}
\label{sec:algorithm}

In this section, we prove that if the FD schema $\mathbf{R}$ has an lhs chain, then there is a polynomial time algorithm for $\textsc{CR-NN}_<\langle \mathbf{R} \rangle$ in the size of the inconsistent dataset, the parameter $k$ and the number of possible labels. Then, we show that when $\mathbf{R}$ consists of a single primary key we can construct an even faster algorithm that runs in {\em linear time} w.r.t the number of tuples, number of labels, and $k$.

For this section, let $D$ be an inconsistent labeled instance and $x$ be the test point. Assume w.l.o.g. that $\mathcal{Y} = \{1,2, \dots, m\}$ and let $n$ be the number of tuples in $D$. We assume that the points in $D$ are already in strict order w.r.t.\ their distance from $x$: $t_1 < t_2 < \dots < t_n$.

\subsection{An Algorithm for Certifiable Robustness}\label{sec:algo1}

Note that in the following analysis we fix an FD schema $\mathbf{R}$ with an lhs chain. The algorithm first constructs a repair $I$ by choosing greedily points using the given ordering as long as they do not conflict with each other. This step can be implemented in time $O(n)$ by, say, building a hash map per FD which maps for each tuple the value of the LHS attribute(s) to the value of the RHS attribute(s).
Suppose w.l.o.g. that $\mathcal{L}_I(x) = 1$. 

As a second step, for every label $\ell \in \{2, \dots, m\}$, we will attempt to construct a repair $I'$ of $D$ such that the number of $\ell$-labeled points is at least as many as the number of 1-labeled points in the $k$-neighborhood of $x$. Such a repair will be a witness that some other label is possible for $x$, hence $x$ is not certifiably robust.

It will be helpful now to define the following terms for a subinstance $I \subseteq D$,  $\tau \in \{1, \dots, n\}$,  and a label $\ell \in \mathcal{Y}$:
\begin{align*}
 \mathcal{N}^{\leq}_{\tau}(I) & = \{ t_j \in I \mid j \leq \tau \} \\
C^\leq_\tau(\ell,I) & = |\{ t \in \mathcal{N}^{\leq}_{\tau}(I) \mid L(t) = \ell\}|
\end{align*}
%

We are now ready to present the core component of our algorithm. This component will be executed for every label $ \ell > 1$ and $\tau \in \{1, \dots, n\}$. Thus, it will run $O(|\mathcal{Y}| \cdot n)$ times. Define the following quantity for a subinstance $J \subseteq D$, an FD set $\Delta$, and $i$ where $0 \leq i \leq k$:
$$M_i[J, \Delta] = \max \{ C^\leq_\tau(\ell,K) - C^\leq_\tau(1,K) \mid K \in I_\Delta(J) \text{ s.t. } |\mathcal{N}^{\leq}_{\tau}(K)| = i \} .$$
Here for simplicity we adopt a slight abuse of notation where, although $M_i[J, \Delta]$ depends on $\tau$, $\tau$ is not explicitly shown in the notation $M_i[J, \Delta]$. If there is no repair $K$ for $J$ such that $|\mathcal{N}^{\leq}_{\tau}(K)| = i$, we define $ M_i[J,\Delta] = - \infty$. The key observation is that if $M_k[D,\Sigma] \geq 0$ then $\ell$ is a possible label for $x$ and hence robustness is falsified. The algorithm computes this quantity using a combination of dynamic programming and recursion on the structure of the FD set. 

\smallskip
\introparagraph{The Recursive Algorithm}
Given $J \subseteq D$ and a set of FDs $\Delta$, we want to compute $M_i[J,\Delta]$ for every $i=0, \dots, k$. First, we remove from $\Delta$ any trivial FDs. Then we distinguish three cases:
\begin{description}
    \item[Base Case:] {\em the set of FDs is empty.} In this case, $I_\Delta(J) = \{J\}$. For every $i \neq |\mathcal{N}^{\leq}_{\tau}(J)|$, $M_i[J,\Delta] = -\infty$. For  $i = |\mathcal{N}^{\leq}_{\tau}(J)|$ (if $|\mathcal{N}^{\leq}_{\tau}(J)| \leq k$), we have $M_i[J,\Delta] = C^\leq_\tau(\ell,J) - C^\leq_\tau(1,J)$, so we can compute this in a straightforward way.
    
    \item[Consensus FD:]  {\em there exists an FD $\emptyset \rightarrow A$.} In this case, we recursively call the algorithm to compute $M_i[\sigma_{A=a}(J),\Delta-A]$ for every $a \in \pi_A(J)$. Then, for every $i = 0, \dots, k$: 
    $$M_i[J,\Delta] = \max_{a \in \pi_A(J)} M_i[\sigma_{A=a}(J),\Delta-A]$$
    
    \item[Common Attribute:] {\em there exists a common attribute $A$ in the lhs of all FDs.} In this case, we recursively call the algorithm to compute $M_i[\sigma_{A=a}(J),\Delta-A]$ for every $a \in \pi_A(J)$. Let $S = \pi_A(J) = \{a_1, \dots, a_{|S|} \}$. Then, for every $i = 0, \dots, k$: 
    $$M_i[J,\Delta] = \max_{\sum_{a \in S} i_a=i} \sum_{a \in S} M_{i_a}[\sigma_{A=a}(J),\Delta-A]$$
We next discuss how to do the above computation using dynamic programming. We can view $M_{i_a}[\sigma_{A=a}(J),\Delta-A]$ as a matrix with rows indexed by value $a$ of attributes $A$ and columns indexed by $i_a$ with $0 \leq i_a \leq k$. The task is to pick one entry from each row so that the sum of entries is maximized and the column indices of entries sum to $i$. 
The dynamic programming computes an $|S| \times (k+1)$ matrix $\mathcal{M}$ where the $(i,j)$-entry represents the maximum of $\sum_{u=1}^{i}M_{i_u}[\sigma_{A=a_u}(J),\Delta-A]$ such that $\sum_{u=1}^{i}i_u=j$ and finally returns the entries $\mathcal{M}[|S|,i]$. 
\end{description}

\begin{algorithm}[ht!]
\SetAlgoLined

 \For{$j=0, \dots, k$}{
 $\mathcal{M}[1,j] \leftarrow M_j[\sigma_{A=a_1}(J),\Delta-A]$ \; }
  \For{$i=2, \dots, |S|$}{
  \For{$j=0, \dots, k$}{
 $\mathcal{M}[i,j] \leftarrow \max_{u} \{ \mathcal{M}[i-1,u]+M_{j-u}[\sigma_{A=a_i}(J), \Delta-A]\}$ \; } }
 \caption{Dynamic Programming}
 \label{Dynamic Programming}
\end{algorithm}

\begin{lemma}\label{lemma:decision_tractable}
The Recursive Algorithm outputs the correct result and runs in polynomial time with respect to the size of $D$, the parameter $k$ and the number of labels $|\mathcal{Y}|$.
\end{lemma}

\begin{proof}
To prove the correctness of the algorithm, we use the observation that if the functional dependencies form an lhs chain, they can be arranged in an order $X_1 \rightarrow Y_1, \dots, X_n \rightarrow Y_n$ such that $X_i \subseteq X_j$ for all $i<j$. Note the 3 cases are mutually exclusive, meaning that at every execution step the algorithm will go into one and only one case. Note also that the algorithm will eventually terminate since step 2 and 3 will eliminate at least one lhs or rhs attribute of an FD and the FD schema $\mathbf{R}$ is fixed. 

We now argue that the algorithm will return the correct result for $\textsc{CR-NN}_<\langle \mathbf{R} \rangle$. The crucial observation is that $M_i[\sigma_{A=a}(J),\Delta-A]$ represents the maximum difference between the number of label $\ell$ and the number of label $1$ within all repairs of $J$, where $A=a$ and the number of tuples with distance to the test point less than $\tau$ is exactly $i$. Thus, the formula correctly computes such maximum difference among all repairs of $J \subseteq D$ by summing all repairs $K \subseteq J$ with fixed value of the attribute $A$ with suitable $i_j$'s. The fact that a repair can be ``reassembled'' by sub-repairs partitioned according to different values of an attribute hinges on the assumption that the FD schema $\mathbf{R}$ has an lhs chain. We conclude that the algorithm will correctly return $M_k[D,\Sigma]$ whose non-negativity is equivalent to non-certifiable-robustness.

Regarding the running time, observe that for every label $\ell$ and every threshold $\tau=1, \dots, n$, the algorithm will be called recursively finitely many times since the FD schema $\mathbf{R}$ is fixed. Furthermore, the dynamic programming in step 3 will run in $O(n\cdot k^2)$ time, since the matrix $\mathcal{M}$ has size $O(n\cdot k)$ and to compute each entry we need $O(k)$ time. Thus for each $\ell$ and $\tau$, the algorithm will run in polynomial time with respect to $n$ and the parameter $k$. Since there are $O(n)$ thresholds, we conclude that the algorithm runs in polynomial time with respect to the size of $D$, the parameter $k$ and the number of labels $|\mathcal{Y}|$.
\end{proof}

\introparagraph{Weighted Majority} The algorithm can also handle the case where we compute the predicted label by weighted majority, where each tuple $t$ is assigned a weight $w_t$. The only thing we need to modify is the definition of $C^\leq_\tau(\ell,I)$, which now becomes $\sum_{ t \in \mathcal{N}^{\leq}_{\tau}(I): L(t) = \ell} w_t$.

\subsection{A Faster Algorithm for Single Primary Key}


The algorithm given in the above section, though in polynomial time, is not very efficient. 
In this section, we give a faster algorithm for certifiable robustness when the FD schema is equivalent to one with a single primary key. 
Recall that in this case we need to pick exactly one tuple from the set of tuples that share the same key. 

As in the previous section, we will first run $k$-NN on an arbitrarily chosen repair to obtain a possible label for $x$ (this part needs only linear time). Without any loss of generality, assume that the predicted label for $x$ is 1. For every target label $\ell \in \{2, \dots, m\}$, we will attempt to construct a repair such that $\ell$ occurs at least as many times as 1 in the $k$-neighborhood of $x$. If such a repair exists, then robustness is falsified.


\begin{figure*}[ht!]
\definecolor{Cblue}{RGB}{1,86,153}
\definecolor{Cyellow}{RGB}{250,192,15}
\definecolor{Corange}{RGB}{243,118,74}
\definecolor{Clightblue}{RGB}{95,198,201}
\definecolor{Cpurple}{RGB}{79,89,100}
\begin{subfigure}[b]{\linewidth}
\centering
\begin{tikzpicture}[scale = 0.8]
\draw node at (0,0) [fill=Corange,circle] {1};

\draw node at (0.75,0) [fill=Cblue,rectangle] {1};

\draw node at (1.5,0) [fill=Corange,circle] {1};

\draw node at (2.5,0) [fill=Cpurple,diamond] {3};

\draw node at (3.75,0) [fill=Cpurple,diamond] {1};

\draw node at (5,0) [fill=Cyellow,star] {1};

\draw node at (6,0) [fill=Cblue,rectangle] {3};

\draw node at (7,0) [fill=Cblue,rectangle] {1};

\draw node at (8,0) [fill=Clightblue,ellipse] {2};

\draw node at (9,0) [fill=Cpurple,diamond] {1};

\draw node at (10,0) [fill=Clightblue,ellipse] {3};

\draw node at (11,0) [fill=Clightblue,ellipse] {3};

\draw node at (12,0) [fill=Cyellow,star] {1};

\draw node at (13,0) [fill=Clightblue,ellipse] {1};

\draw node at (14,0) [fill=Cpurple,diamond] {3};

\draw node at (15.25,0) [fill=Cyellow,star] {3};

\end{tikzpicture}

\caption{Ordering of the tuples in instance $D$}
\label{Linear ordering of instance $D$}
\end{subfigure}

\begin{subfigure}{1.0\linewidth}
\centering
\begin{tikzpicture}[scale = 0.8]
\draw node at (0,0) [fill=white,circle] {};

\draw node at (1.5,0) [fill=Corange,circle] {1};

\draw node at (2.5,0) [fill=Cpurple,diamond] {3};

\draw node at (6,0) [fill=Cblue,rectangle] {3};

\draw node at (7,0) [fill=Cblue,rectangle] {1};

\draw node at (8,0) [fill=Clightblue,ellipse] {2};

\draw node at (14,0) [fill=Cpurple,diamond] {3};

\draw node at (15.25,0) [fill=Cyellow,star] {3};
\end{tikzpicture}
\caption{Result of $\textsc{Prune}(D, 2, 1)$}
\label{Prune(D, 2, 1)}
\end{subfigure}

\begin{subfigure}{1.0\linewidth}
\centering
\begin{tikzpicture}[scale = 0.8]
\draw node at (0,0) [fill=white,circle] {};

\draw node at (1.5,0) [fill=Corange,circle] {1};

\draw node at (2.5,0) [fill=Cpurple,diamond] {3};

\draw node at (6,0) [fill=Cblue,rectangle] {3};

\draw node at (8,0) [fill=Clightblue,ellipse] {2};

\draw node at (10,0) [fill=Clightblue,ellipse] {3};

\draw node at (15.25,0) [fill=Cyellow,star] {3};
\end{tikzpicture}
\caption{Result of $\textsc{Prune}(D, 3, 1)$}
\label{Prune(D, 3, 1)}
\end{subfigure}

\caption{Running example for the single primary key algorithm. Tuples with the same color/shape belong in the same block.}
\label{Running Example}
\end{figure*}
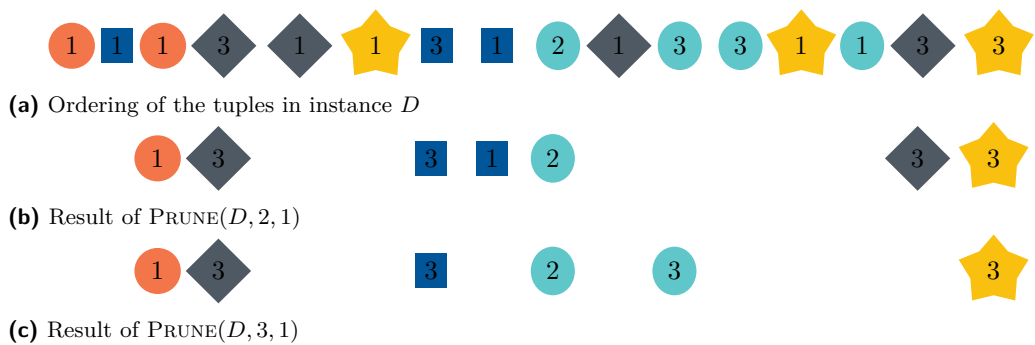

For a tuple $t$, we denote $\mathsf{key}(t)$ to be its key. The {\em block} of a tuple $t$ is the set of tuples with the same key. Further, define $C(\ell,I) = |\{ t \in \mathcal{N}_{k}(x,I) \mid L(t) = \ell \}| $.

Suppose now we want to find a repair $I$ such that $C(\ell_2, I) \geq C(\ell_1, I)$. Define $\textsc{Prune}(D, \ell_2, \ell_1)$ to be the dataset obtained from $D$ if we remove any tuple $t \in D$ such that there exists another tuple $t' \in D$ in the same block with:
\begin{enumerate}
 \item $t < t'$ and $L(t) = \ell_1$; or
 \item $t > t'$ and $L(t') = \ell_2$.
\end{enumerate} 

\begin{lemma} \label{lem:helper}
Let $\ell_1, \ell_2 \in \mathcal{Y}$ and $D^\star = \textsc{Prune}(D, \ell_2, \ell_1)$. Then, there exists a repair $I$ of $D$ s.t. $C(\ell_2, I) \geq C(\ell_1, I)$ if and only if there exists a repair $I'$ of $D^\star$ with $C(\ell_2, I') \geq C(\ell_1, I')$.
\end{lemma}

\begin{proof}
For the proof of Lemma~\ref{lem:helper}, we need the following helper Sublemma.

\begin{sublemma} \label{lemma:h}
Let $t,t' \in D$ such that $t < t'$ and $\key{t} = \key{t'}$. Consider any repair $I$ of $D$ with $t \in I$ and let $I' = (I \setminus \{t\}) \cup \{t'\}$ be another repair of $D$. Then for any label $\ell$:
\begin{align*}
 C(\ell,I') - C(L(t),I') \geq  C(\ell,I) - C(L(t),I). 
 \end{align*}
\end{sublemma}

\begin{proof}[Proof of Sublemma~\ref{lemma:h}]
We distinguish two cases. Suppose $t \notin \mathcal{N}_k(x,I)$. Since $t' > t$, the $k$-neighborhood of $x$ remains the same, and thus the inequality holds with equality. Suppose now that $t \in \mathcal{N}_k(x,I)$. In this case, $t$ is replaced by some other tuple $t''$ in the $k$-neighborhood. If $L(t) = L(t'')$, the inequality holds again with equality. Otherwise, $C(L(t),I') = C(L(t),I) -1$. Since $|C(\ell, I) - C(\ell,I')| \leq 1$, the inequality follows.
\end{proof}

Since $D^\star \subseteq D$, the one direction is straightforward. For the other direction, suppose that there exists a repair $I$ of $D$ s.t. $C(\ell_2, I) \geq C(\ell_1, I)$. We will transform $I$ to a repair $I'$ of $D^\star$ with the same property by constructing a sequence of datasets $(D = )D_0 \supset D_1  \supset  \dots$ such that for every $i$ there exists a repair $I_i$ of $D_i$ with $C(\ell_2, I_i) \geq C(\ell_1, I_i)$. The invariant is true for $i=0$. For some $i >0$, if $ I_i \subseteq D^\star$, then $I_i$ is a repair of $D^\star$ with the desired property. Otherwise, let $t \in I_i \setminus D^\star$. Then, two cases exist:
\begin{itemize}
\item $L(t) = \ell_1$ and there exists a tuple $t' \in D_i$ in the same block as $t$ with $t  < t'$.  Let $D_{i+1}= D_i \setminus \{t\}$ and its repair $I_{i+1}= (I_i \setminus \{t\}) \cup \{t'\}$. Then, by applying Lemma~\ref{lemma:h} with $\ell = \ell_2$, $I = I_i$ and $I' = I_{i+1}$, we obtain $C(\ell_2,I_{i+1}) - C(\ell_1,I_{i+1})  \geq C(\ell_2,I_i) - C(\ell_1,I_i) \geq 0$, where the last inequality comes from the invariant.
\item There exists a tuple $t' \in D_i$ in the same block as $t$ with $t'  < t$ and $L(t') = \ell_2$.  Let $D_{i+1}= D_i \setminus \{t\}$ and its repair $I_{i+1}= (I_i \setminus \{t\}) \cup \{t'\}$. Then, by applying Lemma~\ref{lemma:h} with $\ell = \ell_1$, $I = I_{i+1}$ and $I' = I_{i}$, we obtain $C(\ell_2,I_{i+1}) - C(\ell_1,I_{i+1})  \geq C(\ell_2,I_i) - C(\ell_1,I_i) \geq 0$, where the last inequality comes from the invariant.
\end{itemize}
Finally, note that the sequence must terminate at some point since the datasets $D_i$ strictly decrease in size. 
\end{proof}

Lemma~\ref{lem:helper} tells us that it suffices to consider $D^\star$ instead of $D$. $D^\star$ has the following nice properties:
\begin{itemize}
\item every block has at most one tuple with a label from $\{\ell_1,\ell_2\}$.
\item  any tuple with label in $\{\ell_1,\ell_2\}$ is always the last tuple in its block (i.e. the one furthest away from $x$).
\end{itemize}
The pruning procedure can be implemented in linear time $O(n)$.
Algorithm~\ref{FastScan} \textsc{FastScan} now attempts to find the desired repair. It runs in linear time with respect to the size of $D$ and the number of labels $|\mathcal{Y}|$ and, moreover, its time complexity does not depend on $k$. 

\begin{algorithm}[ht]
\SetAlgoLined

\SetKwInOut{Output}{output}
\KwIn{instance $D$, test point $x$, labels $\ell_1, \ell_2$}
\KwOut{whether there exists repair I s.t. $C(\ell_2,I) \geq C(\ell_1,I)$}

$D \gets \textsc{Prune}(D, \ell_2, \ell_1)$ \;
 $n_1, n_2 \gets 0$ \;
 $\textsf{B}, \textsf{B}^\square  \gets \{\}$ \;
 \For{$i \leftarrow 1$ \KwTo $|D|$} {
 $\textsf{B} \gets \textsf{B} \cup \{\key{t_i}\}$ \;
  \If{$L(t_i) = \ell_2$}{$n_2 \gets n_2 +1$\;}
  \If{$t_i$ is the only tuple of its block and $L(t_i)=\ell_1$}{$n_1 \gets n_1 +1$\;}
  \If{$t_i$ is the last tuple of its block}{$\textsf{B}^\square  \gets \textsf{B}^\square  \cup \{\key{t_i}\}$\;}
  \If{$|\textsf{B}^\square| \leq k \leq |\textsf{B}|$ and  $n_2 \geq n_1$}{\Return true\;}
  }
 \Return false\;
 \caption{\textsc{FastScan}}
 \label{FastScan}
\end{algorithm}

\begin{theorem} \label{thm:fast}
There exists an $O(|\mathcal{Y}|n)$ algorithm for $\textsc{CR-NN}_< \langle \mathbf{R} \rangle$ when the FD schema $\mathbf{R}$ is equivalent to one with a single primary key.
\end{theorem}

\begin{proof}
We first show the correctness of our Algorithm~\ref{FastScan}. The algorithm falsifies robustness if there exists a pair of labels $(\ell, 1)$ for which \textsc{FastScan} returns true.  Suppose that \textsc{FastScan} terminates at iteration $i$. Then, we can construct a repair $I$ of $D^\star$ (and hence of $D$) as follows. For each block in $\textsf{B}^\square$ we pick the last tuple unless its label is $\ell_1$, in which case we arbitrarily pick another tuple in that block. For the blocks in $\textsf{B} \setminus \textsf{B}^\square$, we pick any tuple $ \leq t_i$ for $(k- |\textsf{B}^\square|)$ blocks, and for the remaining blocks in $\textsf{B} \setminus \textsf{B}^\square$ we pick a tuple $> t_i$. For the remaining blocks we make an arbitrary choice. From our construction, $I$ has exactly $k$ tuples $\leq t_i$, which form the $k$-neighborhood. Since we have pruned $D$, all the occurrences of tuples with label $\ell_1, \ell_2$ occur at the last positions of the blocks. Hence, $C(\ell_2, I)- C(\ell_1, I) = n_2 - n_1\geq 0$.


For the opposite direction, suppose that there exists a repair $I$ that falsifies robustness. Then, for some label $\ell$ we must have $C(\ell_2, I)- C(\ell_1, I) \geq 0$.  From Lemma~\ref{lem:helper}, there exists a repair $I'$ of $D^\star$ for which $C(\ell_2, I') \geq C(\ell_1, I')$. Let $t_i$ be the last tuple in the $k$-neighborhood of $I'$. Then, the exit condition will be satisfied for the $i$-th iteration of the loop of \textsc{FastScan} and hence the algorithm will return true.


\smallskip

We now discuss the time complexity. First, we spend $O(n)$ to find a possible label for $x$. The algorithm runs \textsc{FastScan} $(m-1)$ times, for every pair of labels $(\ell, 1)$ where $\ell = 2, \dots, m$.  Finally, we need to argue that each iteration of the loop in \textsc{FastScan} requires $O(1)$ time. This can be achieved by implementing \textsf{B} and $\textsf{B}^\square$ as hash-sets with constant-time insertions. Checking whether a tuple is the last one in a block can also be done in $O(1)$ time by performing a preprocessing linear pass that marks all the last tuples of every block.
\end{proof}

When $|\mathcal{Y}|=2$, the algorithm essentially reduces to the MinMax algorithm in~\cite{KarlasLWGC0020}. For $|\mathcal{Y}| \geq 3$ it outperforms the SortScan algorithm~\cite{KarlasLWGC0020}, since the latter algorithm has an exponential dependence on $|\mathcal{Y}|$ and $k$. Our algorithm also can deal with the case where two tuples in the same block have different labels, which is not something the MinMax and SortScan algorithms can handle.

\begin{example}
We now illustrate our algorithm by a simple example where $k=3$ and $\mathcal{Y} = \{1,2,3\}$. Figure~\ref{Linear ordering of instance $D$} represents an inconsistent instance $D$, where nodes with the same shape/color are in the same block. Their distances to a given test point are increasing from left to right. A repair that chooses the first two tuples assigns label 1 to $x$, hence 1 is a possible label. Figures~\ref{Prune(D, 2, 1)} and~\ref{Prune(D, 3, 1)} illustrate the pruned instances \textsc{Prune}(D, 2, 1) and \textsc{Prune}(D, 3, 1), respectively. Take Figure~\ref{Prune(D, 3, 1)} for example: when \textsc{FastScan} reaches the iteration where $i=3$, we have $n_2 =2, n_1 = 1$, $|B^{\square}| = 3$ and $|B| = 3$, so $|B^{\square}| =  k = |B|$ and $n_2 \geq n_1$. Indeed, by choosing the first, second, third, and last two tuples in Figure~\ref{Prune(D, 3, 1)}, we see that label 1 is not robust (against label 3).
\end{example}

\section{Hardness Result}
\label{sec:hardness}

In this section, we establish the main intractability result.

\begin{theorem}\label{thm:hardness-1}
Let $\mathbf{R}$ be an FD schema that is not equivalent to any FD schema with an lhs chain. Then, the problem $\textsc{CR-NN}_p\langle \mathbf{R},1\rangle$ is \textsf{coNP-complete} for any $p>1$.
\end{theorem}

\begin{proof}
The \textsf{coNP} membership of the problem $\textsc{CR-NN}_p\langle \mathbf{R},1\rangle$ follows from the observation that if one is not certainly robust, then it can be checked efficiently two given repairs (certificate) indeed lead to two different prediction labels. To prove this hardness result, we will describe a reduction from the SAT-3-restricted problem, inspired by the construction of~\cite{YG80} for the edge dominating set problem. In this variant of SAT, each clause has at most three literals, while every variable occurs two times and its negation once. 

Let $\phi$ be a SAT-3-restricted formula. Suppose that $\phi$ has $m$ clauses $C_1, C_2, \dots, C_m$ and $n$ variables $x_1, x_2, \dots, x_n$. Our construction consists of three steps. 

\smallskip
\noindent {\bf Step 1:}
In the first step, we construct a directed labeled graph $G = (V,E)$ with labels in $\{0,1\}$:
\begin{itemize}
    \item The set of vertices $V = \{C_i \bigcup x^{k}_j \bigcup y^{k}_j \text{ where } 1 \leq i \leq m, 0 \leq k \leq 2 \text{ and } 1\leq j \leq n \}$.
    \item For each clause $C_i$, where $i=1, \dots, m$, we add the following labeled edge:
    $$(C_i^+, C_i^-) \rightarrow 0$$
    That is, we add the directed edge which points from $C_i^+$ to $C_i^-$ to the set of edges $E$ and label it as $0$.
    \item Suppose that variable $x_j$, where $j=1, \dots, n$, appears positive in clauses $C_\kappa, C_\lambda$, and negative in clause $C_\mu$. Then, we introduce the following labeled edges:
    \allowdisplaybreaks
    \begin{align*} 
& (C_\kappa^+, x_j^0), (y_j^0, x_j^0) \rightarrow 1 \\
& (C_\lambda^+, x_j^1), (y_j^1,x_j^1)  \rightarrow 1 \\
& (x_j^2, C_\mu^-), (x_j^2, y_j^2)  \rightarrow 1\\
& (y_j^0, y_j^2), (y_j^1, y_j^2)   \rightarrow 0
\end{align*}
\end{itemize}
Figure~\ref{fig:gadget-hardness} shows the above construction.
Note that $G$ is a directed bipartite graph, since no vertex has both incoming and outgoing edges. Hence, one can equivalently view each maximal matching of $G$ as a subset repair of an instance with FD schema $(R(A,B),\{A \rightarrow B, B \rightarrow A\})$ and vice versa (attributes $A$ and $B$ correspond to the two sides of the bipartite graph).

	\begin{figure}[!ht]
	    \centering 
    \begin{tikzpicture}[->,>=stealth,auto=left, semithick,scale=2,vnode/.style={black,inner sep=2pt,scale=1},el/.style = {inner sep=3/2pt}]
      \node[vnode] (c0) at (1,0) {$y^0_j$};
      \node[vnode] (p1) at (1,4/3) {$C_\kappa^+$};
       \node[vnode] (p2) at (3,4/3) {$C_\lambda^+$};
       \node[vnode] (d0) at (1,2/3) {$x_j^0$};
      \node[vnode] (c2) at (2,0) {$y_j^2$};
      \node[vnode] (d2) at (2,2/3) {$x_j^2$};
      \node[vnode] (d1) at (3,2/3) {$x_j^1$};
      \node[vnode] (c1) at (3,0) {$y_j^1$};
      \node[vnode] (m) at (2,4/3) {$C_\mu^-$};

       \path[->] (c0) edge node[el] {0} (c2); 
       \path[->] (c1) edge node[el] {0} (c2); 
       \path[->,densely dotted] (d2) edge node[el] {$1$} (c2);  
       \path[->,densely dashed] (c1) edge node[el] {$1$} (d1);  
	\path[->,densely dashed] (c0) edge node[el] {$1$} (d0);  
	\path[->,densely dashed] (d2) edge node[el] {$1$} (m);  
	\path[->,densely dotted] (p1) edge node[el] {$1$} (d0);
	\path[->,densely dotted] (p2) edge node[el] {$1$} (d1);    

    \end{tikzpicture}	   
	    \caption{Variable gadget for the hardness reduction from the SAT-3-restricted problem.}
	    \label{fig:gadget-hardness}
	\end{figure}
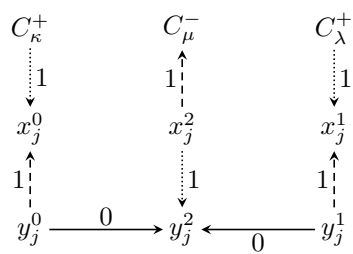

\begin{sublemma} \label{sub:1}
$\phi$ is satisfiable if and only if there exists a maximal matching for $G$ that includes only edges with label 1.
\end{sublemma}

\begin{proof}[Proof of Sublemma~\ref{sub:1}]
$\Rightarrow$  Assume that the variable assignment $\psi$ makes $\phi$ satisfiable. Fix any order of variables $x_1 \dots, x_n$. We form a set of edges $M$ as follows. For any variable $x_j$ visited in the above order, we distinguish two cases:
\begin{itemize} 
\item $\psi(x_j) = \textsf{true}$: we pick $(x_{j}^{2}, y_{j}^{2})$.  If $C_{\kappa}^{+}$ is not yet matched, pick $(C_{\kappa}^{+}, x_{j}^{0})$, otherwise pick $(y_{j}^{0}, x_{j}^{0})$. Similarly for $C_{\lambda}^{+}$. 
\item $\psi(x_j) = \textsf{false}$: pick $(y_{j}^{0}, x_{j}^{0})$ and $(y_{j}^{1}, x_{j}^{1})$. If $C_{\mu}^{-}$ is not yet matched, pick $(x_{j}^{2}, C_{\mu}^{-})$, otherwise pick $(x_{j}^{2}, y_{j}^{2})$. 
\end{itemize}

By construction, $M$ contains only edges with label 1. 

\smallskip

{\em Claim 1: $M$ is a matching.} Since $x_j, y_j$ occur only in a variable gadget, they will have at most one adjacent edge from $M$. By construction, each clause $C_\kappa^+, C_\mu^-$ will also have at most one adjacent edge from $M$.

\smallskip

{\em Claim 2: $M$ is a maximal matching.} First, observe that by our construction, all $x_j^0, x_j^1, x_j^2$ are matched for any $j=1, \dots, n$. Second, notice that if $\psi(x_j) = \textsf{true}$, the edge $(x_{j}^{2}, y_{j}^{2})$ will be chosen; if $\psi(x_j) = \textsf{false}$, the edges $(y_{j}^{0}, x_{j}^{0})$ and $(y_{j}^{1}, x_{j}^{1})$ will be chosen. Thus, the edges $(y_j^0, y_j^2), (y_j^1, y_j^2)$ can not be added to $M$.
Finally, consider the edge $(C_i^+, C_i^-)$ corresponding to clause $C_i$.  If there exists a positive literal which satisfies $C_{i}$, then consider the earliest $x_{j}$ in the linear order of variables. By construction, $(C_i^+, x_j^\nu)$ is in the matching, where $\nu \in \{0,1\}$. Otherwise, $C_{i}$ is satisfied by a negative literal: consider the earliest such $x_k$ in the linear order.  Then $(x_{k}^2, C_{i}^{-})$ is in $M$.
In either case, $(C_i^+, C_i^-)$ cannot be added to increase the size of the matching.

$\Leftarrow$ Assume a maximal matching $M$ that avoids 0-labeled edges. For every variable $x_j$, if $M$ contains $(x_j^2, y_j^2)$, we assign $\psi(x_j) = \textsf{true}$, otherwise $\psi(x_j) = \textsf{false}$. We claim that $\psi$ is a satisfying assignment for $\phi$. Indeed, take any clause $C_i$. Since $(C_i^+, C_i^-) \notin M$, there  exists some edge in $M$ that conflicts with it. If this edge is of the form $(C_i^+, x_j^\nu)$,  then it must be that $(x_j^2, y_j^2) \in M$. But then $\psi(x_j) = \textsf{true}$, and since $x_j$ occurs positively in $C_i$ the clause is satisfied. If this edge is of the form $(x_j^2, C_i^-)$, then $(x_j^2, y_j^2) \notin M$. Thus, $\psi(x_j) = \textsf{false}$, and since $x_j$ occurs negatively in $C_i$ it is again satisfied. 
\end{proof}

\smallskip
\noindent {\bf Step 2:} A maximal matching of $G$ can be viewed equivalently as a repair of a labeled instance $D_0$ with FD schema $\mathbf{S} = (R(A,B),\{A \rightarrow B, B \rightarrow A\})$. In the second step, we will transform the instance $D_0$ of $\mathbf{S}$ to a labeled instance $D_1$ of the target FD schema $\mathbf{R}$. We will do this using the concept of {\em fact-wise reductions}. A fact-wise reduction from $\mathbf{S}$ to $\mathbf{R}$ is a function $\Pi$ that maps a tuple from an instance of $\mathbf{S}$ to a tuple of an instance of $\mathbf{R}$ such that $(i)$ it is injective, $(ii)$ it preserves consistency and inconsistency (i.e.\ a tuple in $D_0$ violates $\mathbf{S}$ if and only if the corresponding tuple in $D_1$ violates $\mathbf{R}$), and $(iii)$ it can be computed in polynomial time. In fact, we will use exactly the same fact-wise reduction as the one used in~\cite{LKW21} to reduce an instance in $\mathbf{S}$ to one in $\mathbf{R}$, where $\mathbf{R}$ is not equivalent to an FD schema with an lhs chain. It will be necessary to present this reduction in detail, since its structure will be needed for the third step of our reduction.

W.l.o.g., we can assume the FD schema is minimal. Since it does not have an lhs chain, there are two FDs $X \rightarrow A$ and $X' \rightarrow A'$ such that $X \subsetneq X'$ and $X' \subsetneq X$. Let $\oplus$ be a fresh constant. We map $t = R(u,v)$ with label $\ell$ to a tuple $\Pi(t)$ with label also $\ell$ such that:
\begin{align*}
  \Pi(t)[A_i] = \begin{cases} 
   \oplus &\mbox{if } A_i \in (X \cap X')^{+,\Sigma}  \\
    u  &\mbox{if } A_i \in X \setminus (X \cap X')^{+,\Sigma}  \\
    v  &\mbox{if } A_i \in X' \setminus (X \cap X')^{+,\Sigma}  \\
    (u,v) & \mbox{otherwise.}
\end{cases}
\end{align*}

Here, $X^{+,\Sigma}$ denotes the closure of an attribute set $X$ w.r.t.\ the FD set $\Sigma$.
By~\cite{LKW21} we know that $\Pi$ is a fact-wise reduction. Let $D_1$ be the resulting instance of $\mathbf{R}$. 

\begin{sublemma} \label{sub:2}
$\phi$ is satisfiable if and only if there exists a repair for $D_1$ in $\mathbf{R}$ that includes only tuples with label 1.
\end{sublemma}

\begin{proof}[Proof of Sublemma~\ref{sub:2}]
This follows from Sublemma~\ref{sub:1} and the fact that $\Pi$ is a fact-wise reduction.
\end{proof}

\smallskip
\noindent {\bf Step 3:} In the last step of the reduction, we will present an encoding $ \llbracket \cdot \rrbracket$ that embeds the values of the attributes in $D_1$ to values in $\mathbb{N}$, hence allowing us to compute distances with the $p$-norm. The resulting tuples will also be labeled from $\mathcal{Y} = \{0,1\}$.

Let $\alpha = d \cdot (2m+8n)$, where $d$ is the number of attributes. First, let $\llbracket \oplus \rrbracket = 0$. For a vertex $u \in V$, let
\begin{align*}
   \llbracket u \rrbracket = \begin{cases} 
   i &\mbox{if } u = C_i^+   \\
   m+i &\mbox{if } u =  C_i^-   \\
   2m + 3j+\nu &\mbox{if } u = y_j^\nu  \\
   \alpha + 3j+\nu &\mbox{if } u = x_j^\nu 
\end{cases}
\end{align*}
It is easy to see that the above embedding is injective, meaning that if $\llbracket u \rrbracket = \llbracket v \rrbracket$ then $u=v$.
As for the edges, consider any ordering $e_1, e_2, \dots$ and simply let $\llbracket e_i \rrbracket = i$. Note that the number of edges in $G$ is $|E| = 8n + 2m$. This embedding is also injective. Let $D_2 = \llbracket D_1 \rrbracket $ denote the instance we obtain by encoding every value of $D_1$ as above. Since the encoding is injective, this is also trivially a fact-wise reduction, hence Sublemma~\ref{sub:2} holds for $D_2$ as well. 

\begin{sublemma} \label{sub:3}
$\phi$ is satisfiable if and only if $x= R(0,0,\dots, 0)$ has no certain label in $D_2$.
\end{sublemma}

\begin{proof}[Proof of Sublemma~\ref{sub:3}]
We first need the following two claims.

\smallskip
\noindent {\bf Claim 1:} {\em Any tuple with label 0 is closer to $x$ than any tuple with label 1.} Indeed, a tuple has label 1 if and only if it contains in an attribute a value of the form $\llbracket x_j^\nu \rrbracket$. Hence, any tuple with label 1 has distance $> \alpha$ from $x$. On the other hand, each attribute in a 0-labeled tuple has value at most $2m+8n$. Hence, the distance from any tuple with label 0 is bounded by $d^{1/p} \cdot (2m+8n) \leq d \cdot (2m+8n)$.

\smallskip
\noindent {\bf Claim 2:} {\em 0 is a possible label for $x$.} Indeed, the tuple corresponding to the edge $(C_1^+, C_1^-)$ is the closest one to $x$ and has label 0. Hence, any repair that includes this tuple will assign the label 0 to $x$.

\smallskip
$\Rightarrow$ Assume that the variable assignment $\psi$ makes $\phi$ satisfiable. Then, we know that there exists a repair for $D_2$ that avoids any tuple with label 0. This repair will then assign label 1 to $x$, which implies that $x$ is not a certain label since by {\bf Claim 2} 0 is a possible label for $x$.

\smallskip
$\Leftarrow$ Assume a repair that assigns a label 1 to $x$ -- hence making $x$ have no certain label. Since by \textbf{Claim 1} all 0-labeled tuples are closer than the 1-labeled tuples, this means that all tuples in the repair must have label 1. But then, $\phi$ is satisfiable.
\end{proof}

This completes the proof.
\end{proof}

Finally, we  extend the intractability result from $\textsc{CR-NN}_p\langle \mathbf{R},1\rangle$ to $\textsc{CR-NN}_p\langle \mathbf{R},k\rangle$ for any integer $k \geq 1$.

\begin{theorem}\label{thm:hardness-k}
Let $\mathbf{R}$ be an FD schema that is not equivalent to any FD schema with an lhs chain. Then, for any integer $k \geq 1$, $\textsc{CR-NN}_p\langle \mathbf{R},k\rangle$ is \textsf{coNP-hard} for any $p>1$.
\end{theorem}

\begin{proof}
To show this hardness result, we will show a reduction from $\textsc{CR-NN}_p\langle \mathbf{R},1\rangle$ to $\textsc{CR-NN}_p\langle \mathbf{R},k\rangle$, $k> 1$, for any FD schema $\mathbf{R}$ and binary labels $\mathcal{Y} = \{0,1\}$. We will also assume that we restrict the inputs such that the test point $x$ does not coincide with any existing tuple in the dataset (this assumption does not affect the hardness claim).

Let $D$ the input labeled instance with values in $\mathbb{N}$, and $x \notin D$ be the test point. Assume w.l.o.g. that $x = (0, \dots, 0)$; otherwise, we can shift the coordinates so that distances are maintained and $x$ becomes the origin. Find the tuple $t_0$ in $D$ that is closest to $x$ and let $d_0 = f(x,t_0)$ (this can be done in polynomial time). Since $x \neq t_0$, $d_0 > 0$. W.l.o.g. assume $ L(t_0) = 0$. Let $\lambda = d^{1/p}k / d_0$. 

We construct an instance $D'$ as follows. First, we map each tuple $R(a_1, \dots, a_d)$ in $D$ to $R(\lambda \cdot a_1, \dots, \lambda \cdot a_d)$. Second, we add the tuples $T = \{ R(1,1, \dots, 1), R(2, \dots, 2), R(k-1, \dots, k-1)\}$ with corresponding labels $0,1,0, \dots$.  

By construction, every new point in $T$ has distance at most $d^{1/p}(k-1)$ from $x$. On the other hand, all the other points have distance at least $d^{1/p}k$. Moreover, the tuples in $T$ do not conflict with the existing tuples or with each other. From this, we obtain that the $k$-neighborhood of $x$ in every repair of $D'$ contains $T$.

We now claim that {\em $x$ is certifiably robust in $D$ for 1-NN classification if and only if it is certifiably robust in $D'$ for $k$-NN classification}.

\smallskip
\noindent $\Rightarrow$ Suppose that $x$ is certifiably robust in $D$ for 1-NN classification. Since $0$ is a possible label for $x$, this means that for every repair of $D$, the closest point $t$ has label $0$. But then every repair of $D'$ will be $T$ plus one point that has label $0$ in the $k$-neighborhood of $x$, so the number of $0$-labeled tuples will be at least $\lceil (k-1)/2 \rceil +1$, which consists a majority.

\smallskip 
\noindent $\Leftarrow$  Suppose that $x$ is not certifiably robust in $D$ for 1-NN classification. Since 0 is a possible label for $x$, this means that there exists a repair of $D$ such that the closest tuple $t$ has label 1. But then the corresponding repair of $D'$ will have $T$ plus one point that has label $1$ in the $k$-neighborhood of $x$, so the number of $1$-labeled tuples will be at least $\lfloor (k-1)/2 \rfloor +1$, which is either majority or tie. 
\end{proof}

\section{Optimal Repairs Revisited}
\label{sec:forbidden}

In this section, we investigate the complexity landscape of a related problem to certifying robustness for $k$-NN classification, which may be of independent interest.  In~\cite{LKR18}, the authors studied the \textsc{Opt-Repair} problem: each tuple $t$ is associated with a positive weight $w_t \geq 0$, and we want to find the optimal subset repair that removes the set of tuples with the smallest total weight. Note that this is equivalent to finding the repair with the largest total weight. 

Here, we consider the symmetric problem, \textsc{Min-Repair}: we want to find {\em the subset repair that has tuples with the smallest total weight}. In this case, we interpret the tuple weights as a measure of how "wrong" we think a tuple is. We can parametrize this problem with a given FD schema $\mathbf{R}$, as in $\textsc{Min-Repair} \langle \mathbf{R} \rangle$.
\textsc{Min-Repair} captures as a special case the following problem, denoted as $\frep{\mathbf{R}}$: given an inconsistent instance $D$ over $\mathbf{R}$ and a subinstance $S \subseteq D$, does there exist a subset repair $I \subseteq D$ such that $I \cap S = \emptyset$?

\begin{lemma}
There exists a many-one polynomial time reduction from $\frep{\mathbf{R}}$ to $\textsc{Min-Repair} \langle \mathbf{R} \rangle$.
\end{lemma}

\begin{proof}
One can set the weight of any tuple in $S$ to be 1, otherwise 0. Then, there exists a repair that avoids the forbidden set $S$ if and only if there exists a repair with total weight equal to $0$. 
\end{proof}

From the hardness proof of Theorem~\ref{thm:hardness-1}, we immediately obtain the following intractability result.

\begin{theorem}
Let $\mathbf{R}$ be an FD schema that is not equivalent to any FD schema with an lhs chain. Then, $\frep{\mathbf{R}}$ is \textsf{NP-hard}. As a result, $\textsc{Min-Repair} \langle \mathbf{R} \rangle$ is also \textsf{NP-hard}.
\end{theorem}

It turns out that the forbidden set repair problem is directly connected with certifying robustness for $1$-NN classification.

\begin{lemma} \label{lem:forb}
There exists a many-one polynomial time reduction from $\frep{\mathbf{R}}$ to the complement of $\textsc{CR-NN}_{<}\langle \mathbf{R},1 \rangle$.
\end{lemma}

\begin{proof}
Let $D$, $S\subseteq D$ be the inputs to the \textsc{Forbidden-Repair} problem. We will construct a labeled instance for the classification problem using only two labels, $\mathcal{Y} = \{0,1\}$. 
The input instance is exactly $D$.
For labeling, if $t \in S$ then $L(t)=0$, otherwise $L(t)=1$. Finally, we construct an ordering of the tuples in $D$ such that $t < t'$ whenever $t \in S$, $t' \in D \setminus S$.

We first claim that any repair of $D$ that avoids $S$ is a repair that assigns a label of $1$ to $x$. Indeed, the $1$-neighborhood of $x$ for such a repair will consist of a tuple with label 1 by construction. On the other hand, any repair that includes a tuple from $S$ assigns label $0$ to $x$.
Since there exists at least one repair includes a tuple from $S$, there exists no repair that avoids the forbidden set $S$ if and only if $x$ is certifiably robust (with label 0). 
\end{proof}

The above reduction gives a polynomial time algorithm for $\textsc{Forbidden-Repair} \langle \mathbf{R} \rangle$ whenever $\mathbf{R}$ is equivalent to an FD schema with an lhs chain. We now present Algorithm~\ref{Min-Rep} that works for the more general \textsc{Min-Repair} problem.

\begin{algorithm}[h]
\SetAlgoLined

\If{$\Sigma$ is trivial}{\Return $D$}
remove trivial FDs from $\Sigma$ \;
\If{$\Sigma$ has a common lhs attribute $A$}
	{\Return $\cup_{a \in \pi_A(D)} \textsc{Min-Rep}(\Sigma-A,\sigma_{A=a}(D))$}
\If{$\Sigma$ has a consensus FD $\emptyset \rightarrow A$}
	{$m \gets \arg \min_{a \in \pi_A(D)} w(\textsc{Min-Rep}(\Sigma-A,\sigma_{A=a}(D)))$ \;
	\Return $\textsc{Min-Rep}(\Sigma-A,\sigma_{A=m}(D))$}

 \caption{$\textsc{Min-Rep}(\Sigma,D)$}
 \label{Min-Rep}
\end{algorithm}

The algorithm works similarly to the \textsf{OptSRepair} algorithm in~\cite{LKR18} with two differences. First, we only need to consider the cases where the FD schema has a common lhs or a consensus FD (i.e., FD with an empty lhs). Second, in the case of the consensus FD, where we partition the instance, we take the repair that has the {\em minimum} total weight instead of the largest one.

\begin{theorem}
Let $\mathbf{R}$ be an FD schema that is equivalent to an FD schema with an lhs chain. Then, $\textsc{Min-Repair} \langle \mathbf{R} \rangle$ is in \textsf{P}.
\end{theorem}

The dichotomy we obtain for \textsc{Min-Repair} coincides with the one we obtain for $\textsc{CR-NN}$. However, it is different from the dichotomy observed for the \textsc{Opt-Repair} problem in~\cite{LKR18}; in fact, every FD schema that admits a polynomial time algorithm for $\textsc{Min-Repair}$ also admits a polynomial time algorithm for \textsc{Opt-Repair}, but not vice versa. Specifically, the FD schema $(R(A,B), \{A \rightarrow B, B \rightarrow A\})$ is hard for $\textsc{Min-Repair}$, but tractable for $\textsc{Opt-Repair}$. The reason is that finding a maximum-weight (maximal) matching in a bipartite graph is polynomially solvable, but finding a minimum-weight maximal matching is \textsf{NP-hard}.

The next lemma might be of independent interest.
\begin{lemma}
There exists a Cook reduction from $\textsc{CR-NN}_<\langle \mathbf{R},1 \rangle$ to $\textsc{Forbidden-Repair}\langle \mathbf{R}\rangle$.
\end{lemma} 

\begin{proof}
We are given an instance $D$, a test point $x$. Let $t_1, t_2, \dots, t_n$ be the tuples in increasing order of their distance to $x$. Observe that we can always choose a repair that includes the closest tuple $t_1$. Hence, the label $\ell = L(t_1)$ is a possible label for $x$. To show that $x$ is certifiably robust, it suffices to prove that none of the labels in $\mathcal{Y} \setminus \{\ell\}$ are possible. 

Let $\ell' \in \mathcal{Y} \setminus \{\ell\}$.
For every tuple $t_i$ such that $L(t_i) = \ell'$, we construct an input to $\textsc{Forbidden-Repair}\langle \mathbf{R}\rangle$ as follows. First, construct $D'_i$ to be the instance where we have removed from $D$ the tuple $t_i$ along with any tuples that conflict with $t_i$. Then, the input to  $\textsc{Forbidden-Repair}$ is the instance $D'_i$ along with the forbidden set $\{t_1, \dots, t_{i-1} \} \cap D'_i$. In other words we ask: is there a repair that includes $t_i$ but avoids all the other tuples that are closer to $x$?
It is now easy to observe that $\ell'$ is a possible label for $x$ -- and thus $x$ is not certifiably robust -- if and only if at least one instance (out of the at most $n$ instances) for $\textsc{Forbidden-Repair}$ returns true. 
\end{proof}

\section{Certifiable Robustness by Counting}
\label{sec:counting}

In this section, we consider the counting version of certifiable robustness for $k$-NN classifiers. The counting problem of certifiable robustness asks the following: among all possible repairs, how many will predict label $\ell$ for a fixed $\ell \in \mathcal{Y}$?


We show that the counting problem still remains in polynomial time if the FD set $\mathbf{R}$ is equivalent to an lhs chain by generalizing the algorithm in Section~\ref{sec:algo1}. Formally, the class of counting problems that can be computed in polynomial time is called \textsf{FP}. 

\begin{theorem} \label{thm:counting_tractable}
If the FD schema $\mathbf{R}$ is equivalent to some FD schema with an lhs chain, then the counting problem $\textsc{\#CR-NN}_<\langle \mathbf{R} \rangle$ is in  \textsf{FP}. 
\end{theorem}

We show now how to generalize the algorithm in Section~\ref{sec:algo1} to perform counting. 
Let $x$ be the test point and $\mathcal{Y} = \{ 1, ,2 \dots, m\}$. 
It suffices to show how to count for the label $\ell=1$. Recall the definition of $\mathcal{N}^{\leq}_{\tau}(x, I)$ and $C^\leq_\tau(\ell,I)$. Similarly as in Section~\ref{sec:algo1}, we will compute a high-dimensional matrix $M$ ``layer by layer'' and read off the answer from it.

We now make our key definition. Fix a threshold $\tau > 0$, and define the following quantity for a (possibly inconsistent) subinstance $J \subseteq D$ and integers $i, c_2, c_3, \dots, c_m$, where $0 \leq i \leq k$ and $-k \leq c_j \leq k$ for all $j \in \{2, 3, \dots, m\}$:

\begin{equation*}
\begin{split}
M_i[J, c_2, \dots, c_m, \Delta] = \{ \# K \mid K \in I_\Delta(J) \text{ s.t. } |\mathcal{N}^{\leq}_{\tau}(x, K)| = i \text{ and} \\ C^\leq_\tau(j,K) - C^\leq_\tau(1,K) = c_j \ \forall j \in \{2, 3, \dots, m\}\}.
\end{split}
\end{equation*}

For simplicity of notation, let ${\mathbf{c}}$ denote the vector $(c_2, c_3, \dots, c_m)$ with the understanding that $\mathbf{c}_j$ could represent any new vector $(c_{2_j}, c_{3_j}, \dots, c_{m_j})$. Sometimes we might suppress the FD set $\Delta$ to write $M_i[J,{\mathbf{c}}]$ when the context is clear. 
The interpretation for the entry $M_i[J,{\mathbf{c}}]$ is that it records the number of repairs in $J$ with $i$ many tuples which are among $\tau$-th closest to $x$ such that the differences between the number of $1$-tuples and the number of $j$-tuples are exactly $c_j$, $j \in \{2, 3, \dots, m\}$. If there is no such  $K \in I_\Sigma(J)$, we define $M_i[J,{\mathbf{c}}]=0$. The algorithm computes the entries of $M$ by a combination of dynamic programming and recursion on the FD schema $\mathbf{R}$. Note that after computing $M$, the answer to $\textsc{\#CR-NN}_<\langle \mathbf{R} \rangle$ is exactly the sum of all entries $M_k[D, {\mathbf{c}}]$ where $c_j \geq 1$ for all $j \in \{2,3,\dots,m \}$. The algorithm has three disjoint cases when computing $M_i[J,{\mathbf{c}}]$:

\begin{description}

	\item[Base Case:] {\em the set of FDs is empty.} In this case, $M_i[J,{\mathbf{c}}] = 0$ unless $|\mathcal{N}^{\leq}_{\tau}(x, I)| = i \leq k$ and $C^\leq_\tau(j,J) - C^\leq_\tau(\ell,J) = c_j$ for all $j \in \{2, 3, \dots, m\}$, in which case the entry is 1. This step clearly can be computed efficiently.
    
	 \item[Consensus FD:]  {\em there exists an FD $\emptyset \rightarrow A$.} In this case, we recursively call the algorithm to compute $M_i[\sigma_{A=a}(J),{\mathbf{c}},\Delta-A]$ for every value $a \in \pi_A(J)$. Then, we calculate $M_i[J,{\mathbf{c}}] = \sum_{a \in \pi_A(J)} M_i[\sigma_{A=a}(J),{\mathbf{c}}, \Delta-A]$.
 
	\item[Common Attribute:] {\em there exists a common attribute $A$ in the lhs of all FDs.} In this case, we recursively call the algorithm to compute $M_j[\sigma_{A=a}(J), {\mathbf{c}_j},\Delta-A]$ for every value $a \in \pi_{A}(J)$ and all $j,{\mathbf{c}_j}$ such that $0 \leq j \leq i$ and $0 \leq c_{l_j} \leq c_l$ where $2 \leq l \leq m$. Let $S = \pi_A(J) = \{a_1, \dots, a_{|S|} \}$. We then compute 
$$M_i[J,{\mathbf{c}}] = \sum \sum\limits_{j=1}^{|S|} M_{i_j}[\sigma_{A=a_j}(J), {\mathbf{c}_j},\Delta-A]$$ 
where the first summation is over all possible solutions of $i_j$'s and $c_{l_j}$'s such that $\sum_{j=1}^{|S|}i_j=i$ and $\sum_{j=1}^{|S|}c_{l_j}=c_l$ for all $l \in \{2,3,\dots,m\}$. We now show how to compute $M_i[J,{\mathbf{c}}]$ by a direct generalization of Algorithm~\ref{Dynamic Programming}. The dynamic programming computes a $(m+1)$-dimensional matrix $\mathcal{M}$ where its $(p,q,{\mathbf{c}_i})$-entry represents the sum $\sum_{j=1}^{p}M_{i_j}[\sigma_{A=a_j}(J), {\mathbf{c}_j}]$ over all possible solutions $i_j$'s and ${\mathbf{c}_j}$'s such that $\sum_{j=1}^{p} i_j = q$ and $\sum_{j=1}^{p}c_{l_j} = c_{l_i}$ for $2 \leq l \leq m$. Note that $\mathcal{M}$ has $|S| \cdot (k+1) \cdot (2k+1)^{m-1} = O(n \cdot k^m)$ entries.
Let $K = \{k, k-1, \dots, -k \}$ and ${\mathbf{c}_i} - {\mathbf{c}_j} := (c_{2_i}-c_{2_j}, \dots, c_{m_i}-c_{m_j})$. We are now ready to present our dynamic programming algorithm:
	
\begin{algorithm}[ht]
\SetAlgoLined

 \For{$j=0, \dots, k \text{ and } {\mathbf{c}} \in K^m$ }{
 $\mathcal{M}_j[1,{\mathbf{c}}] \leftarrow M_j[\sigma_{A=a_1}(J), {\mathbf{c}}, \Delta-A]$ \; }
  \For{$i=2, \dots, |S|$}{
  \For{$j=0, \dots, k \text{ and } {\mathbf{c}} \in K^m$}{
 $\mathcal{M}_j[i, {\mathbf{c}}] \leftarrow \sum\limits_{p, \mathbf{c}_q} \{ \mathcal{M}_p[i-1,\mathbf{c}_q]+M_ {j-p}[\sigma_{A=a_i}(J), \mathbf{c} - \mathbf{c}_q,\Delta-A]\}$\; } }
 \caption{Dynamic Programming (Counting)}
 \label{Dynamic Programming (Counting)}
\end{algorithm}


\end{description}


We now show that the counting algorithm returns correctly and runs in polynomial time with respect to the size of $D$ and the parameter $k$. However, the running time depends exponentially on the number of labels $|\mathcal{Y}|$. An open question remains whether this exponential dependency is essential.

\begin{proof}[Proof of Theorem~\ref{thm:counting_tractable}]
The correctness of the counting algorithm follows similarly as in the proof of Lemma~\ref{lemma:decision_tractable}. We now argue that the counting algorithm runs in polynomial time. For every label $\ell$ and every threshold $\tau$, the algorithm will be called recursively finitely many times since the FD schema $\mathbf{R}$ is fixed. Furthermore, the dynamic programming in step 3 will run in $O(n\cdot k^{2m})$, where $n=|D|$, since the matrix $\mathcal{M}$ has size $O(n \cdot k^m)$ and to compute each entry we need $O(k^m)$ time. Thus for each $\ell$ and $\tau$, the algorithm will run in polynomial with respect to $n$ and the parameter $k$. Since there are $O(n)$ boundary thresholds, we conclude that the algorithm runs in polynomial time with respect to the size of $D$ and the parameter $k$.
\end{proof}

The hardness part for counting is an immediate corollary with the previous result in~\cite{LKW21}.

\begin{theorem} \label{thm:sharp}
If the FD schema $\mathbf{R}$ is not equivalent to some FD schema with an lhs chain, then $\textsc{\#CR-NN}_<\langle \mathbf{R} \rangle$ is \textsf{\#P-complete}.
\end{theorem}

\begin{proof}
By Theorem 3.2 in~\cite{LKW21}, it is \textsf{\#P-hard} to count the number of repairs for the FD schema $\mathbf{R}$. Now, given any instance $D$, we can pick any ordering of the points and assign the same label $\ell$ to every tuple. Then, the number of repairs that predict label $\ell$ is the same as the total number of repairs.
\end{proof}

The approximate counting of certifying robustness of $k$-NN classifiers is also hard. 

\begin{theorem} \label{thm:approx}
If the FD schema $\mathbf{R}$ is not equivalent to some FD schema with an lhs chain, then $\textsc{\#CR-NN}_<\langle \mathbf{R} \rangle \equiv_{AP} \textsc{\#SAT}$.
\end{theorem}

Here, $\equiv_{AP}$ refers to the equivalency up to approximation-preserving reductions~\cite{DyerGGJ03}.

\begin{proof}
By the same fact-wise reduction in Lemma~\ref{sub:2} and labelling in Theorem~\ref{thm:sharp}, we have $\textsc{\#MaximalBIS} \leq \textsc{\#SRep} \leq \textsc{\#CR-NN}_<\langle \mathbf{R} \rangle$ where $\textsc{\#MaximalBIS}$ is the problem of counting the number of maximal independent sets in a bipartite graph and $\textsc{\#SRep}$ is the problem of counting the number of subset repairs of an inconsistent instance. By Theorem 1 in~\cite{GoldbergGL16}, $\textsc{\#MaximalBIS} \equiv_{\text {AP }} \textsc{\#SAT}$ and thus we obtain  $\textsc{\#CR-NN}_<\langle \mathbf{R} \rangle \equiv_{AP} \textsc{\#SAT}$.
\end{proof}

\section{Other Uncertainty Models}
\label{sec:other}

In this section, we study the complexity of certifying robustness for $k$-NNs under three simple uncertainty models: $?$-sets, or-sets, and Codd tables. We show that for these models we can certify robustness in polynomial time. Throughout the section, we fix the relational schema to be $R(A_1, \dots, A_d)$.

\introparagraph{?-Sets with Size Constraints} For a given instance $D$ over the schema, we mark an uncertain subset $D_?$ of the tuples in $D$. Then, for a positive integer $m \geq 1$, we define the set of possible worlds as:
$$ \mathcal{I}_? = \{I \mid D \setminus D_? \subseteq I \subseteq D, |D \setminus I| \leq m \}. $$

In other words, we can construct a possible world by removing any -- but at most $m$ -- tuples from $D_?$. When $m = |D_?|$, this definition captures the notion of $?$-tables as defined in~\cite{SBHW06}. When $D_? = D$, it captures {\em data poisoning} scenarios, where the goal is to protect against attacks that can poison up to $m$ tuples of the training set.

For this setting, we can show that certifiable robustness for $k$-NNs can be computed in almost linear time in the size of the input.

\begin{lemma} \label{lem:?set}
We can certify robustness in $\mathcal{I}_?$ for $k$-NNs in time $O((|\mathcal{Y}| +\log n) \cdot n)$ where $n$ is the size of the dataset.
\end{lemma}

\begin{proof}
For the sake of simplicity, let $\mathcal{Y} = \{1, 2, \dots \}$. 

Let $x$ be the test point. As a first step, we order the tuples in increasing order of $f(x,t)$ -- this can be done in time $O(n \log n)$. We assume w.l.o.g. that all distances are distinct. Let the resulting order be $t_1, t_2, \dots, t_n$.

Since it always holds that $D \in \mathcal{I}_?$, we can fist compute the predicted label of $x$ in $D$ in time $O(k)$.  W.l.o.g. assume that $\mathcal{L}_D(x) = 1$.
For every label $\ell > 1$, the algorithm will now try to construct a possible world where $\ell$ occurs at least as many times as 1 in the $k$-neighborhood of $x$. 
To make the algorithm exposition easier, define for every tuple $t \in D_?$ a priority value $\rho(t)$ as follows.
\begin{align*}
\rho(t) =  \begin{cases}
    2, & \text{if } L(t) = \ell \\
    0, & \text{if } L(t) = 1 \\
    1, & \text{otherwise.}
    \end{cases}
\end{align*}

\begin{algorithm}[ht!]
\SetAlgoLined

\SetKwInOut{Output}{output}
\KwIn{test point $x$}
\KwOut{is $k$-NN  certifiably robust in $\mathcal{I}_?$}

$J \gets \{ t_1, \dots, t_k \} $ \;
$\Delta \gets 0$ \;
 \For{$i = k+1$ \KwTo $n$}{
 $\Delta \gets \Delta + 1$ \;
 \If{$|\{t \in J \mid L(t) = \ell\}| \geq |\{t \in J \mid L(t) = 1\}|$}{\Return \textsf{true}\;}
 \If{$J \cap D_? = \emptyset$ or $\Delta > m$}{\Return \textsf{false}\;}
 $J \gets J \cup \{ t_i \}$ \;
 remove from $J$ the tuple $\arg \min_{t \in J \cap D_?} \{\rho(t)\}$ \;
}
 \caption{Certifiable Robustness for $?$-Sets}
 \label{algo:q-sets}
\end{algorithm}

We now present our Algorithm~\ref{algo:q-sets}. Intuitively, it iterates over the tuples in order of proximity to $x$. For each tuple $t_i$, it attempts to construct the most promising $k$-neighborhood that includes tuples from $\{t_1, \dots, t_i\}$. Each loop in the algorithm can be executed in time $O(1)$. Indeed, we can implement this by keeping three sets with tuples depending on their labels, where we can do insertion and deletion in constant time. 

Note that the running time of Algorithm~\ref{algo:q-sets} is independent of $k,m$ and $|D_?|$, but it does depend linearly on the number of labels.
\end{proof}


\introparagraph{Or-Sets} In this uncertain model, each attribute value of a tuple is an {\em or-set} consisting of finite values (e.g., $\langle 2, 5, 10 \rangle$). Each possible world in $\mathcal{I}_{or}$ is formed by choosing exactly one value from each or-set, independent of the choices across all other or-sets. We can express $\mathcal{I}_{or}$ as subset repairs for the following schema: add to $R$ an extra attribute \textsf{id}. Then, assign a unique value for this attribute to each tuple $t$, and create a tuple for each "realization" of this tuple with the same \textsf{id}. This will increase the input size of the problem, but by at most a polynomial factor (since $d$ is fixed). Moreover, the FD schema is clearly equivalent to an lhs chain; in fact, this is the single primary key case. As a consequence, we have the following proposition. 

\begin{proposition}
We can certify robustness in $\mathcal{I}_{or}$ for $k$-NNs in \textsf{P}.
\end{proposition}


\introparagraph{Codd tables}
In a {\em Codd table}, a missing value is represented as \textsf{Null} paired with a domain \textsf{Dom} from which that value can be chosen. A repair is any possible completion of the table. The first observation is that, by adding a new identifier attribute \textsf{ID}, we can think of a Codd table as an inconsistent instance with a single primary key, where each block has a consistent label (i.e., every tuple in the block has the same label). However,  if \textsf{Dom} is given as an infinite interval, then Algorithm~\ref{FastScan} does not apply directly since it may not be possible to write all tuple completions in an increasing order (there are uncountably many). Indeed, in~\cite{KarlasLWGC0020} Karlas et al. consider only the situation where \textsf{Dom} is given as a finite discrete set.

 Formally, the distance between a tuple $t$ in the Codd table and a test point $x$ is given by the function $f(x,t) = g_t(y_1,y_2, \dots, y_n)$ where the $y_i$'s are the missing entries in $t$. In order to be able to certify robustness, we need to assume that the minimum and maximum of this function can be computed efficiently. This is a mild assumption, since for example for the $p$-norm the minimum and maximum is achieved when each summand is minimized or maximized respectively.

Now, we observe that since every block has the same label, we can modify Algorithm~\ref{FastScan} so that every resulting repair consists of only the first or the last tuple in a block without changing its correctness properties. With this observation in hand, we can now simply define the extremal set $S = \{\min f(x,t) \ \cup \  \max f(x,t) \mid t \in D\}$ consisting of the minimum and maximum of each block and then run Algorithm~\ref{FastScan} on $S$. Since $S$ is at most twice the size of $S$, this implies a linear time algorithm (w.r.t.\ the number of tuples, labels, and $k$) for certifiable robustness on Codd tables. We have shown the following proposition.
\begin{proposition}
We can certify robustness in a Codd table for $k$-NNs in linear time even if the domain \textsf{Dom} is given as an infinite interval.
\end{proposition}
\section{Conclusion}

In this paper, we study the complexity of certifiable robustness for $k$-NN classification under FD constraints. We show a dichotomy in the complexity of the problem depending on the structure of the FDs, and we prove that the same dichotomy condition also holds for the counting version of the problem. We envision this work to be a first step towards the long-term goal of investigating the complexity of certifying robustness for other widely used classification algorithms, such as decision trees, Naive Bayes classifiers and linear classifiers.

\newpage
\bibliography{References}




\end{document}